# Improved Image Classification with Token Fusion

**Keong-Hun Choi[1], Jin-Woo Kim[1,2], Yao Wang[1], Jong-Eun Ha[1]**

[1]Seoul National University of Science and Technology
[2]Technology Research Lab, Ways1 Inc
gyounghun6612@gmail.com, kjwkch@{seoultech.ac.kr, ways1.com}, wangyao@seoultech.ac.kr, jeha@seoultech.ac.kr

**Abstract**

In this paper, we propose a method using the fusion of CNN and transformer structure to improve image classification performance. In the case of CNN, information about a local area on an image can be extracted well, but there is a limit to the extraction of global information. On the other hand, the transformer has an advantage in relatively global extraction, but has a disadvantage in that it requires a lot of memory for local feature value extraction. In the case of an image, it is converted into a feature map through CNN, and each feature map's pixel is considered a token. At the same time, the image is divided into patch areas and then fused with the transformer method that views them as tokens. For the fusion of tokens with two different characteristics, we propose three methods: (1) late token fusion with parallel structure (2) early token fusion (3) token fusion in a layer by layer. In an experiment using ImageNet 1K, the proposed method shows the best classification performance.

## Introduction

After the advent of deep learning, convolutional neural networks (CNNs) have become dominant approaches to image classification, object detection, and semantic segmentation (Szegedy et al 2016; He et al. 2016; Tan and Le 2019; Sandler et al. 2018; Simonyan and Zisserman 2014). Recently, transformers (Yaswani et al. 2017; Devlin et al. 2018) that use various attention mechanisms have become mainstream for natural language processing (NLP) tasks. Applying transformers in the vision domain also shows promising results in multiple tasks, including image classification (Touvron et al. 2020; Dosovitskiy et al. 2020), object detection (Carion et al. 2020; Zhu et al. 2020), and semantic segmentation (Zheng et al. 2020).

Vision transformer (ViT) (Dosovitskiy et al. 2020) firstly showed that an image patch could be regarded as a word token in NLP. ViT for low-level vision (Chen et al. 2020) and SETR (Zheng et al. 2020) for semantic segmentation adopt the design of the ViT.

This paper proposes a method that integrates two types of tokens for image classification. The first type of token is derived from an image patch like the ViT (Dosovitskiy et al. 2020). The second type of token is derived from feature vectors from the feature maps of ResNet (He et al. 2016). Models for various types of fusing two tokens are presented, and experimental results show the feasibility of the proposed algorithm.

## Related Works

CNNs have become mainstream in various applications in computer vision. It has the form of deep neural networks where convolutional layers are stacked in a serial or parallel fashion. Though they showed promising results, they still had difficulties extracting global features due to the limited size of convolutional kernels. Meanwhile, transformers based on self-attention and cross-attention show an advantage in extracting global features compared to the CNNs.

### Convolutional Neural Networks

LeCun et al. (LeCun et al. 1998) proposed the first standard CNN for handwritten character recognition. AlexNet (Krizhevsky, Sutskever, and Hinton 2012) showed a dramatic performance improvement on large-scale image classification tasks (Deng et al. 2009), and it has opened a full-blown prelude in deep learning.

Soon after, various types of newly designed structures based on CNNs such as VGG (Simonyan and Zisserman 2014), GoogleNet (Szegedy et al. 2015), Inception (Szegedy et al. 2016), and ResNet (He et al. 2016) have appeared. ResNet proposed a method to configure a more profound layer with residual connections.

Before the full-scale use of attention in Transformers, some approaches (Wang et al. 2017; Hu, Shen, and Sun 2018; Hu et al. 2018) have used attention mechanisms in CNNs. Wang et al. (Wang et al. 2017) used attention modules sequentially between the intermediate stages of deep residual networks by stacking them. SENet (Hu, Shen, and Sun 2018) and GENet (Hu et al. 2018) use adaptive channel-wise feature responses leveraging interdependencies among channels. NLNet (Wang et al. 2018) introduced self-attention into neural networks, which guarantees reflecting long-range dependencies by pairwise interactions across all spatial positions.

Research has been done on downsizing the CNNs with the tradeoff between accuracy and efficiency (Han et al. 2020; Iandola et al. 2016; Ma et al. 2018). Neural architecture search (NAS) was used to find an efficient structure for CNNs with reduced computation time in MobileNets (Sandler et al. 2018; Howard et al. 38) and EfficientNets (Tan and Le 2019).

**Vision Transformers**

After remarkable achievements by transformers in natural language processing (NLP) (Vaswani et al. 2017; Devlin et al. 2018), many researches (Chu et al. 2021; Touvron et al. 2020; Wang et al. 2021; Liu et al. 2021; Yuan et al. 2021; Dosovitskiy et al. 2020; Han et al. 2021; Wu et al. 2021; Guo et al. 2021; Li et al. 2022; Tang et al. 2021; Tang et al. 2021) have focused on introducing transformer-like structures to various vision tasks.

The pioneering work by ViT (Dosovitskiy et al. 2020) proposed a method to use transformers in NLP on images by considering an image patch as a word token. ViT (Dosovitskiy et al. 2020) used a large private dataset JFT-300M (Sun et al. 2017) to train the model. DeiT (Touvron et al. 2020) proposed an algorithm to train ViT in a data-efficient manner using ImageNet-1K (Deng et al. 2009). In T2T-ViT (Yuan et al. 2021), visual tokens are embedded by recursive aggregation of neighboring tokens into one token. TNT (Han et al. 2021) uses the inner and outer transformer block to represent an image's patch-level and pixel-level characteristics.

PVT (Wang et al. 2021) applied a pyramid structure onto ViT (Dosovitskiy et al. 2020) to use multi-scale feature maps, which is adequate for dense pixel-level prediction tasks. In CPVT (Chu et al. 2021) and CvT (Wu et al. 2021), an algorithm that integrates CNNs and transformers using a convolutional projection into a transformer block is proposed. CMT (Guo et al. 2022) further extends the integration of CNNs and transformers by investigating different components such as shortcut and normalization functions.

Integrating CNNs and transformers focus on model design by intermixing CNN layers and attention blocks. In this paper, we treat them independently to fuse them effectively.

## Method

The proposed method considers the feature map through CNN as a token and also fuses the process of using image patches used in ViT as tokens. CNN has the advantage of extracting local features on the image according to the size of the kernel. On the other hand, it is difficult to reflect the global feature value considering the entire image area. If the kernel size is increased, it is possible to extract global feature values, but there is a difficulty due to a memory problem. In the case of Transformers, it is possible to use image patches as tokens to reflect the global relationship between patches. However, it is challenging to extract local feature values within the patch. Also, a memory problem follows if the patch size is made small.

The proposed method uses a feature value map passed through a particular layer through CNN. Let the feature value at each pixel position in the feature value map be regarded as a token. In the case of a feature value map of the corresponding location, it can be seen as a result of reflecting local feature values through the kernel. Also, like ViT, after dividing the image into patches, use the patches as tokens. The proposed method uses these two different tokens. When combining two tokens, various combinations are possible depending on the difference in the token combination position and the number of tokens. In this paper, we analyze the results according to these combinations through experiments and try to find the configuration that provides the best results.

Figure 1 shows an overview of the proposed method. The part marked with a dotted line is a part that can be selected. Token by CNN and token by ViT are structured to combine before transformer, within, or after the transformer. The right side of Figure 1 shows a commonly used structure in the Transformer.

The Transformer uses 1D sequence of token embeddings as input. A 2D image $\mathbf{x} \in \mathbb{R}^{H \times W \times C}$ is reshaped into a sequence of flattened patches $\mathbf{x}_p \in \mathbb{R}^{N \times (P^2 \cdot C)}$. $(H, W)$ is the height and width of an image, $C$ is the number of channels. The size of an image patch is $(P, P)$ and the number of patches is $N = HW/P^2$. The $i$-th patch $\mathbf{x}_p^i$ is flattened and mapped to $D$ dimensions by a trainable linear projection. The output of linear projection is denoted as the patch embedding. The patch embedding is added with positional encoding $\mathbf{E}_{pos}$ and we denote it as a token, which is used as the input of the Transformer.

$$\mathbf{t}_0 = [\mathbf{x}_p^1 \mathbf{E};\ \mathbf{x}_p^2 \mathbf{E};\ \cdots;\ \mathbf{x}_p^N \mathbf{E}] + \mathbf{E}_{pos},\ \mathbf{E} \in \mathbb{R}^{(P^2 \cdot C) \times D}, \mathbf{E}_{pos} \in \mathbb{R}^{(N \times D)} \quad (1)$$

We treat feature maps by the ResNet-101 model (He et al. 2016) as images. Each pixel on a feature map is considered a token in Equation (1) like ViT (Dosovitskiy et al. 2020).

We use two types of tokens. One is derived from an image patch like ViT (Dosovitskiy et al. 2020). The other is derived from features of the ResNet-101 model (He et al. 2016). These two tokens are integrated into various forms in Figure 1. We do not use a class token in the ViT (Dosovitskiy et al. 2020) because we use two types of tokens. We use all tokens for a classification task. Tokens are processed like the ViT (Dosovitskiy et al. 2020).

$$\mathbf{t}'_l = \text{MHSA}(\text{LN}(\mathbf{t}_{l-1})) + \mathbf{t}_{l-1}, l = 1, \cdots, L \quad (2)$$

$$\mathbf{t}_l = \text{MLP}(\text{LN}(\mathbf{t}'_l)) + \mathbf{t}'_l, l = 1, \cdots, L \quad (3)$$

MHSA, MLP, and LN stand for multi-head self-attention, multi-layer perceptron, and layer norm, respectively.

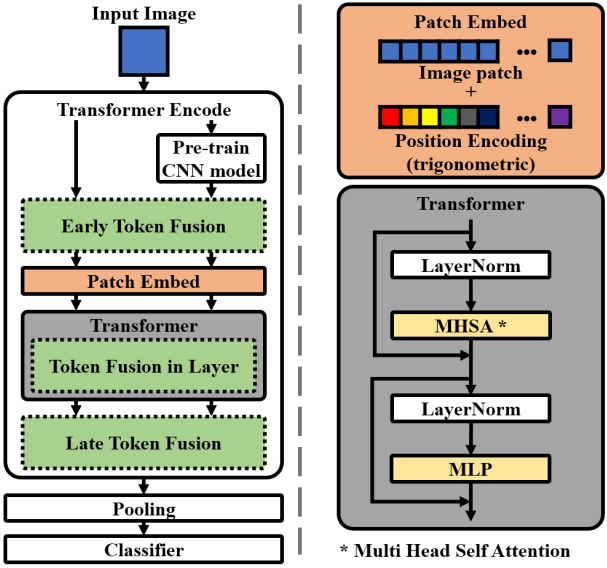

Figure 1: The proposed model with token fusion.

## Method 1: Late Token Fusion with Parallel Structure

The proposed method is to process the token by CNN and the token by image patch in parallel, respectively, and then integrate them. Each token is processed using the encoder structure of the Transformer and has a structure that fuses the processing results.

Figure 2 shows the overall structure. The image processing method by dividing it into patches uses the ViT (Dosovitskiy et al. 2020) structure. Along with this, the feature value map processed by CNN is used in parallel. We use a pre-trained ResNet-101 model (He et al. 2016).

Results from the two transformer encoders have different sizes. To use these fusions, it is necessary to match the size of the feature values. Figure 2(b) shows the detailed process of the late token fusion block. The size of the feature value is expanded four times by using UpConv for the token by ViT. Finally, a token by ViT and a token by CNN are generated as one token using concatenation.

## Method 2: Early Token Fusion

The early token fusion method is processing two tokens before transformer processing. Figure 3 shows the detailed structure of the proposed method. It is a structure that combines the part processed through CNN from the original image and the original image itself.

We use a pre-trained ResNet-101 model (He et al. 2016). Figure 3(b) shows the bride block structure for early token fusion. The result of the corresponding stage of ResNet-101 is generated as one token through concatenation after Up-Conv and 1x1 convolution. Through 5 bridge blocks, you will finally get a unified token of size HXWX18. It should be regarded as an image and processed separately for each patch like ViT.

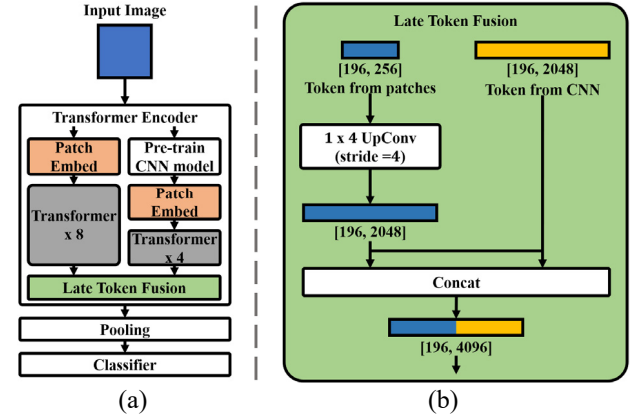

Figure 2: Late token fusion approach by parallel architecture (a) overall model architecture (b) details of late token fusion block

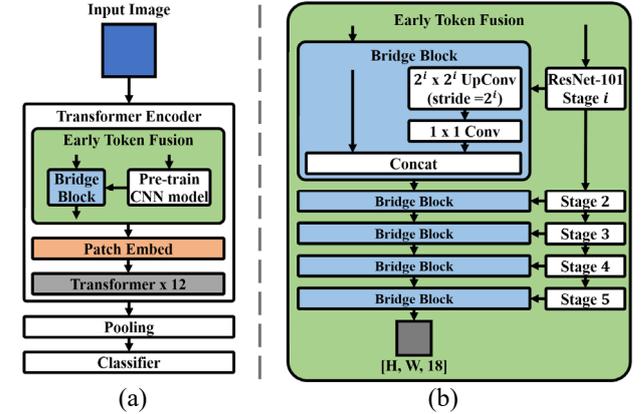

Figure 3: Early token fusion model architecture (a) overall model architecture (b) details of early token fusion block

## Method 3: Token Fusion in a Layer by Layer

The previous two methods are structures that combine tokens before or after transformer processing. The method in this section is a structure that fuses tokens during transformer processing. Figure 4 shows the overall design. It is a structure to apply the mixing block of Figure 4(b) to the stage of ResNet.

The mixing block is composed of a transformer and a CNN, and finally, the results of the two modules are concatenated and integrated. A total of 5 mixing blocks corresponded to the hierarchical structure of ResNet-101. After going through 5 mixing blocks, pass through two transformer blocks.

In the case of the three proposed methods, all models were configured to have a similar size. For this, all three models were configured to have 12 transformer blocks.

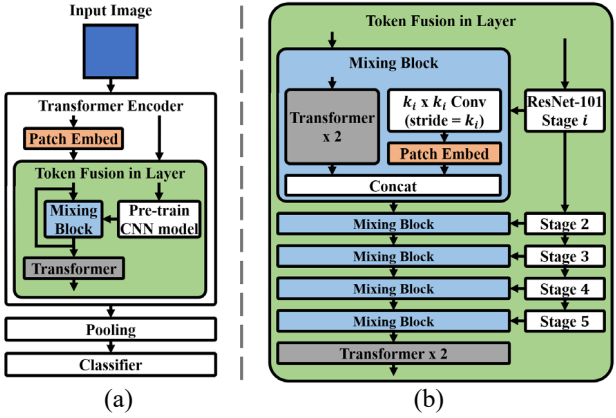

Figure 4: Token fusion in a layer by layer architecture (a) overall model architecture (b) details of token fusion in a layer by layer block

## Image Classification Head

Through the combination of the token by ViT and the token by CNN, it can be seen that information in a form different from the existing transformer structure is extracted. For this reason, the image classification structure was constructed differently from the method using class tokens in the existing ViT (Dosovitskiy et al. 2020). Figure 5 shows the three classification structures used in this paper.

Figure 5(a) is a token-wise pooling classifier, and Figure 5(b) is a channel-wise pooling classifier. Figure 5(c) is a classifier combining token and channel-wise pooling.

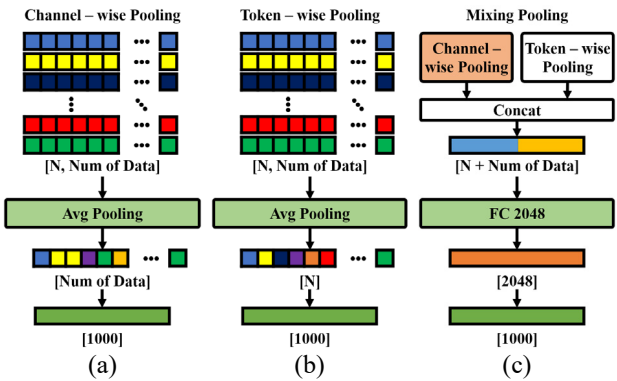

Figure 5: Example of pooling layer (a) Average pooling by data depth (b) Average pooling by token (c) Combined pooling data from each process

## Experimental Results

Experiments are done using i9 10990 CPU and two NVIDIA RTX 3090. Only ImageNet-1K was used for training data. All input images were adjusted to a size of 224(H)X224(W)X3(C). The pre-trained model used in this study was the ResNet-101 model. Therefore, all input images were adjusted to a size of 256(H)X256(W)X3(C), then normalized using only the center 224(H)X224(W)X3(C) and used (mean = [0.485, 0.456, 0.406], std = [0.229, 0.224, 0.225]). Horizontal and vertical flips were applied to the original image for training data enhancement. In addition, the original data was modified and used by adding rotation distortion of up to 15 degrees to the original image.

## Experimental Results using Basic Three Models

Table 1 shows experimental results by the various types of proposed models on the ImageNet-1K. In the case of the early token fusion method, which had the lowest score, it is judged that there is a limit to the improvement of the score because the rest of the model structure is the same as the existing ViT structure except for data combination. The layer-by-layer channel-wise pooling method showed the best results in the experimental results. Through the difference between the corresponding structure and the late token fusion result, we conclude that the related result was obtained by reinforcing the calculation between the data inside the patch and the feature value delivered through the mixing block. In particular, a higher evaluation score could be obtained using a smaller model through this structure.

Table 2 shows comparison results with other algorithms trained only using ImageNet-1K. The best and second-best models are displayed in bold and bold italic, respectively. The same model generally performs better when using a large image as input. The proposed algorithm provides a state-of-the-art result. Also, the proposed algorithm uses 36% fewer parameters than the second-best algorithm of MViTv2-L (Li et al. 2022).

| Method | Classification head type | Evaluation(%) | |
|---|---|---|---|
| | | Acc @1 | Acc @5 |
| Late token fusion | Token-wise pooling | 85.51 | 92.77 |
| | Channel-wise pooling | 85.73 | 91.32 |
| | Mixing | 86.29 | 93.88 |
| Early token fusion | Token-wise pooling | 83.01 | 91.54 |
| | Channel-wise pooling | 84.94 | 93.19 |
| | Mixing | 85.27 | 94.56 |
| Layer by layer | Token-wise pooling | 86.97 | 94.25 |
| | Channel-wise pooling | **87.77** | **95.93** |
| | Mixing | 87.32 | 94.97 |

Table 1: Results using ImageNet-1K by the proposed algorithm.

| Model | Acc @1 (%) | Acc @5 (%) | #Params (M) | Image size |
|---|---|---|---|---|
| CPVT-Ti-GAP (Chu et al. 2021) | 74.9 | - | 6 | $224^2$ |
| DenseNet-169 (Huang et al. 2017) | 76.2 | 93.2 | 14 | $224^2$ |
| ResNet-50 (He et al. 2016) | 76.2 | 92.9 | 25.6 | $224^2$ |
| DeiT-S (Touvron et al. 2020) | 79.8 | - | 22 | $224^2$ |
| DeiT-B (Touvron et al. 2020) | 83.1 | - | 85.8 | $384^2$ |
| EfficientNet-B1 (Tan and Le 2019) | 79.1 | 94.4 | 7.8 | $240^2$ |
| EfficientNet-B6 (Tan and Le 2019) | 84.0 | **96.8** | 43 | $528^2$ |
| ResNeXt-101-64x4d (Xie et al. 2017) | 80.9 | 95.6 | 84 | $224^2$ |
| T2T-ViT-19 (Yuan et al. 2021) | 81.2 | - | 39 | $224^2$ |
| PVT-M (Wang et al. 2021) | 81.2 | - | 44.2 | $224^2$ |
| Swin-T (Liu et al. 2021) | 81.3 | - | 29 | $224^2$ |
| Swin-B (Liu et al. 2021) | 83.3 | - | 88 | $224^2$ |
| CeiT-S (Yuan et al. 2021) | 82.0 | 95.9 | 24.2 | $224^2$ |
| CeiT-S (Yuan et al. 2021) | 83.3 | 96.5 | 24.2 | $384^2$ |
| Twins-SVT-B (Chu et al. 2021) | 83.1 | - | 56 | $224^2$ |
| Twins-SVT-L (Chu et al. 2021) | 83.3 | - | 99.2 | $224^2$ |
| ViT-B/16 (Dosovitskiy et al. 2020) | 77.9 | - | 55.5 | $384^2$ |
| TNT-B (Han et al. 2021) | 82.8 | 96.3 | 65.6 | $224^2$ |
| CvT-21 (Wu et al. 2021) | 83.3 | - | 31.5 | $384^2$ |
| BoTNet-S1-128 (Srinivas et al. 2021) | 83.5 | 96.5 | 75.1 | $256^2$ |
| CMT-S (Guo et al. 2022) | 83.5 | 96.6 | 25.1 | $224^2$ |
| CMT-L (Guo et al. 2022) | 84.8 | **97.1** | 74.7 | $288^2$ |
| MViTv2-B (Li et al. 2022) | 84.4 | - | 52 | $224^2$ |
| MViTv2-L (Li et al. 2022) | *86.0* | - | 218 | $384^2$ |
| Ours | **87.8** | 95.9 | 140 | $224^2$ |

Table 2: Comparison of results with other algorithms using ImageNet-1K.

Table 3 shows the experimental results using ImageNet-22K. The result of our algorithm is by token fusion in a layer by layer model with a channel-wise pooling classification head, which shows the best results among proposed models. The MViT-XL (Li et al. 2022) shows the best result when only ImageNet-22K is used for training. The CoCa (Yu et al. 2022), which gives the best result, uses an additional dataset of JFT-3B for training.

The proposed algorithm shows results closer to the state-of-the-art using ImageNet-22K larger than ImageNet-1K. The proposed algorithm does not give such an improvement that is noticed in other algorithms when using large training datasets such as ImageNet-22K. This requires further investigation.

| Method | | Evaluation(%) | | #Params (M) | Image size |
|---|---|---|---|---|---|
| | | Acc @1 | Acc @5 | | |
| CoCa (Yu et al. 2022) | | 91.00 | - | 2,100 | $576^2$ |
| MViT-XL (Li et al. 2022) | | 88.80 | - | 667 | $512^2$ |
| Ours | ImageNet-22K | 87.95 | 96.08 | 140 | $224^2$ |
| | ImageNet-1K | 87.77 | 95.93 | 140 | $224^2$ |

Table 3: Comparison results using ImageNet-22K (CoCa (Yu et al. 2022) uses additional JFT-3B dataset).

**Experimental Results using Modified Basic Three Models**

In this section, we present experimental results with different model configurations for three types of models.

Figure 6 shows three different configurations in the late token fusion method in Figure 2. UpConv layer was used to match the sizes of two other tokens, and the concatenation layer was used in the combining process. Additional three types of models are configured using copy and element-wise summation.

Figure 7 shows three different configurations in the early token fusion method in Figure 3. The feature values generated through the ResNet-101 model are combined with the original image through a bridge block. We configure three additional models by replacing the UpConv layer inside the bridge block with copy, using only one bridge block, and connecting only the last weight data.

Figure 8 shows a different configuration in the token fusion in a layer by layer in Figure 4. The token combining process is performed through the mixing block of the feature values generated through the ResNet-101 model. The method proceeds without a class token, unlike the existing ViT (Dosovitskiy et al. 2020). In the case of additional experiments, as shown in Figure 8, a class token other than the image patch was added. It was configured to exclude it from the mixing block's combining process.

Table 4 shows experimental results using ImageNet-1K by the modified structure of the proposed model. Results in Table 2 by the basic model show better results than the modified model structure.

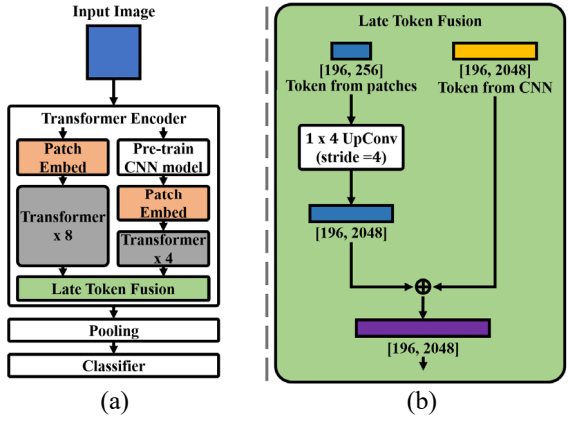

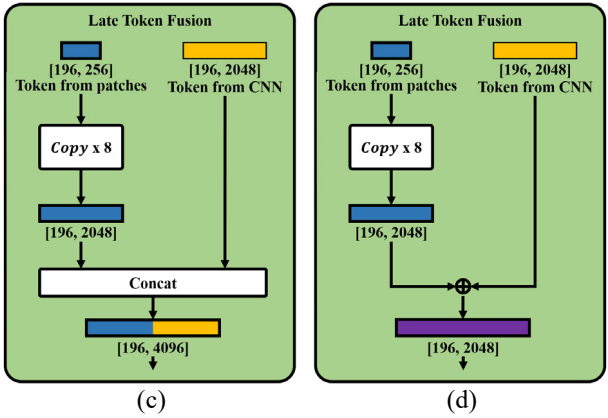

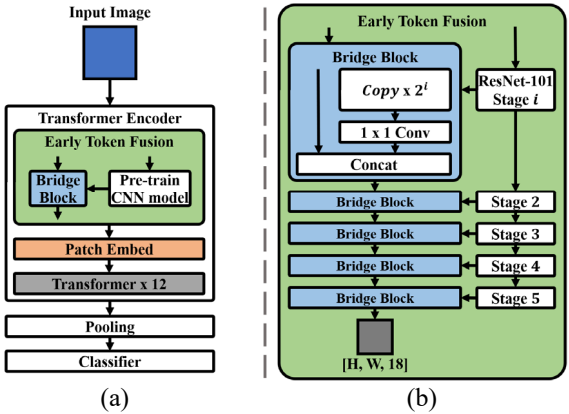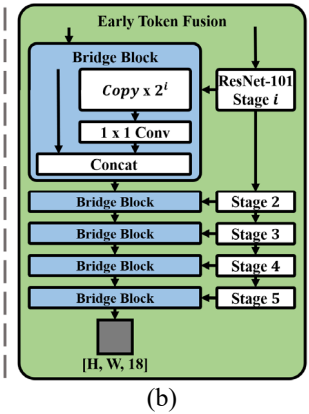

Figure 6: Additional late token fusion approach by parallel architecture (a) overall model architecture (b ~ d) details of late token fusion block (UpConv layer + add, copy + concatenation, copy + add)

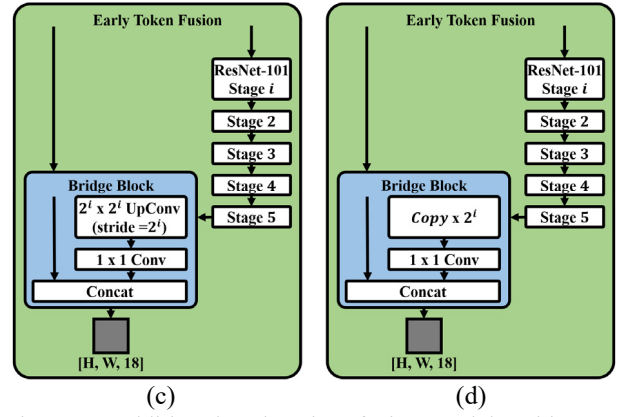

Figure 7: Additional early token fusion model architecture (a) overall model architecture (b ~ d) details of early token fusion block (copy + multi-weight, UpConv + single weight, copy + single weight)

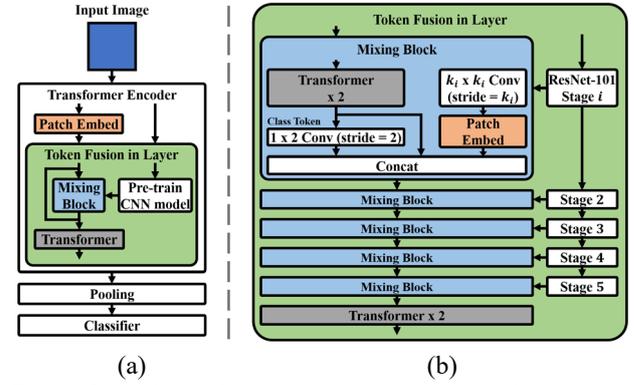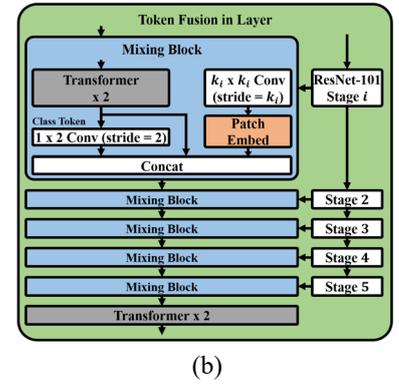

Figure 8: Additional token fusion in a layer by layer architecture (a) overall model architecture (b) details of token fusion in a layer by layer block

| Method | Detail | Classification head type | Evaluation(%) Acc @1 | Acc @5 |
|---|---|---|---|---|
| Late token fusion | Copy + Add | Token-wise | 81.22 | 92.36 |
| | | Channel-wise | 81.64 | 93.01 |
| | Copy + Concat | Token-wise | 81.37 | 90.44 |
| | | Channel-wise | 81.88 | 92.22 |
| | UpConv + Add | Token-wise | 83.35 | 91.54 |
| | | Channel-wise | 83.57 | 92.67 |
| Early token fusion | Copy + Multi | Token-wise | 81.24 | 91.55 |
| | | Channel-wise | 80.41 | 90.55 |
| | Copy + Single | Token-wise | 81.56 | 91.06 |
| | | Channel-wise | 81.54 | 90.69 |
| | UpConv + Single | Token-wise | 82.29 | 90.04 |
| | | Channel-wise | 82.09 | 91.08 |
| Layer by layer | Use class token | Token-wise | 87.14 | 94.54 |
| | | Channel-wise | 87.51 | 95.26 |
| | | Mixing | 87.32 | 94.97 |

Table 4: Results using ImageNet-1K by the proposed algorithm with additional model structure.

## Conclusion

We propose an improved algorithm for image classification with token fusion. Two types are tokens are integrated into the transformer structure. The first type of token is derived from an image patch. The second type of token is derived from feature maps by a CNN. We treat a feature vector per pixel on a feature map as a patch. The CNN has the advantage of extracting local features on an image, while the transformer has the advantage of detecting global features through an attention mechanism. The proposed algorithm efficiently integrates the characteristics of the CNN and transformer. Three models are investigated according to the order and location of token fusion. The proposed algorithm shows the state-of-the-art result for image classification on training only using ImageNet-1K. Further research will extend the proposed model into object detection and semantic segmentation.

## Acknowledgments

This work was supported by a National Research Foundation of Korea (NRF) grant funded by the Korean government (MSIT) (2020R1A2C1013335).